\documentclass[letterpaper, 10pt, conference]{ieeeconf}

\IEEEoverridecommandlockouts
\usepackage[bookmarks=True]{hyperref}
\usepackage{graphicx}
\usepackage{amsmath}
\usepackage{amsfonts}
\usepackage{acronym}

\acrodef{IMU}{internal measurment unit}
\acrodef{GPS}{global positioning system}
\acrodef{SLAM}{simultaneous localization and mapping}
\acrodef{RTK}{real-time kinematics}
\acrodef{CNN}{convolutional neural network}

\title{\LARGE \bf Topological mapping for traversability-aware long-range navigation in off-road terrain}

\author{Jean-Fran\c{c}ois Tremblay, Julie Alhosh, Louis Petit, Faraz Lotfi, Lara Landauro and David Meger
\thanks{
  All authors are with the Center for Intelligent Machines, McGill University, Mont\'eal, Canada.
  We thanks the staff of the Gault Nature Reserve for their support with the experiments.
  Jean-Fran\c{c}ois Tremblay was supported by a Fonds de Recherche du Qu\'ebec - Nature et Technologies Ph.D. scholarship during this research.
  This research was also supported by CMLabs and the National Canadian Robotics Network.
  We thank the staff of McGill's Gault Nature Reserve for their support during the experiments.
}}

\begin{document}

\maketitle

\begin{abstract}
  Autonomous robots navigating in off-road terrain like forests open new opportunities for automation.
  While off-road navigation has been studied, existing work often relies on clearly delineated pathways. 
  We present a method allowing for long-range planning, exploration and low-level control in unknown \textit{off-trail} forest terrain, using vision and GPS only.
  We represent outdoor terrain with a topological map, which is a set of panoramic snapshots connected with edges containing traversability information.
  A novel traversability analysis method is demonstrated, predicting the existence of a safe path towards a target in an image.
  Navigating between nodes is done using goal-conditioned behavior cloning, leveraging the power of a pretrained vision transformer.
  An exploration planner is presented, efficiently covering an unknown off-road area with unknown traversability using a frontiers-based approach.
  The approach is successfully deployed to autonomously explore two 400 m\textsuperscript{2} forest sites unseen during training, in difficult conditions for navigation.
\end{abstract}

\section{Introduction}

Off-road robots have potential applications in forestry \cite{jfr}, search and rescue operations \cite{searchnrescue} and planetary exploration \cite{opportunity}.
However, safely navigating on arbitrarly off-road terrain is challenging due to the unpredictability of the terrain \cite{icra}.
Indeed, terrain can be steep, slippery and full of obstacles.
Dealing with these difficulties, either when controlling the robot or when planning a path ahead, requires dealing with imperfect terrain perception and traversability challenges.
Occupancy is generally a poor measure of traversability for ground vehicles, especially in forests, where vegetation can be trivial to drive through and is often inevitable.
Conversly, free space can be unsafe to drive through, like water or mud.

Here, we propose a method and system for control and long-range planning, enabling the robot to autonomously explore unknown forest environments.
The resulting map, fully covering an area of choice, can then be used for traversability-aware point-to-point navigation.
Our method relies on behavior cloning and topological mapping.
The former negates the need for modeling the complex physics of driving through the forest, like flexible obstacles and terrain, and has been shown to scale to complex driving scenarios \cite{Codevilla_2019_ICCV}.
The latter negates the need for full \ac{SLAM}, dense reconstruction of the environment, allowing us to simply plan using a set of panoramic snapshots of the environment.
Beyond this, topological maps scale well to large environments compared to dense maps. 

Work on on-road and indoor navigation have often considered global planning that requires long-scale decisions but much work on off-road driving have had a short horizon or simple goal-conditioned focus \cite{meng2023terrainnet, badgr}.
In contrast, our system has the ability to plan a path around large obstacles beyond the field of vision, like large fallen trees, streams and such.
The system drives fully off-trail, unlike previous work focusing on a dirt or rocky man-made path \cite{travisicra}.

Our contributions are as follow:
\begin{enumerate}
  \item a visual traversability learning approach, leveraging human expertise, which predicts the existence of a safe path towards a goal in the image,
  \item a topological mapping and exploration framework, where edges in the graph represent traversability as decided by the above model,
  \item a behavior cloning controller capable of navigating very rough terrain, full of slippage, obstacles and tight spaces,
  \item an extensive real-world evaluation in unseen forest terrain, totaling 800 m\textsuperscript{2} of explored terrain and 700 m driven in extremely challeging conditions.
\end{enumerate}

\section{Related works}

\paragraph{Off-road self driving} TerrainNet \cite{meng2023terrainnet} is a method for off-road environment representation and navigation.
It does not consider planning beyond its visual horizon, and requires a dense 3D map to construct its top-down view of the environment.

Manderson et al. \cite{travisicra} present a method for driving on smooth-terrain in a off-road-like environment.
The reward is based on the smoothness of the terrain as measured by an \ac{IMU}.
However, this is not always a viable alternative as rough terrain cannot be avoided in some scenario.
Here, we do not aim to driving on smooth paths, but a general solution for control and planning in off-trail scenarios.

In \cite{frey2023fast}, the authors present a traversability analysis approach which uses a combination of regression and anomaly detection.
This work uses the difference between target velocity and real velocity as a proxy for traversability, and uses this to build a dataset for regression.
For regions where data collection is impossible, they use anomaly detection to detect out-of-distribution region and avoid them.
Rather than having a proxy measure of traversability, like the difference between target and real velocity \cite{frey2023fast}, we rely on human expertise to label traversability and learn from it.

\cite{pmlr-v164-shaban22a,hosseinpoor2021traversability} present a semantic classification approach for navigation in off-road environments.
However, generally speaking, this class of methods is not applicable for off-trail forested environment where there is not a one-to-one mapping between terrain class and traversability.
As an illustration, in \cite{pmlr-v164-shaban22a}, there is a clear human-made path and merely detecting that path is enough to establish where the robot can go.
Compared to \cite{pmlr-v164-shaban22a, hosseinpoor2021traversability} we do not frame the problem as segmentation, which is heavy to label.
Instead, we learn to predict the existence of a traversable path between nodes in our map.



\cite{Baril2021KilometerscaleAN} presents a Teach-and-Repeat approach to driving in subarctic forests.
Teach-and-Repeat allows for repeating a taught trajectory in different times and conditions.
Here we aim to go beyond that by enabling the system to autonomously drive on paths where no demonstrations or teaching phase happened.

\paragraph{Topological mapping}
Dudek et al. \cite{greggraph} present an early work on topological mapping for exploring unknown environments.
Their method assumes no metric or localization method is available, instead relying on markers for place recognition.

\cite{fox} presents an integrated method for local control and planning in topological maps.
However, they do not have to consider traversability contraints as they only work in indoor environments with flat floors.
Generally, existing work on topological mapping focuses on different problems -- mainly loop-closure and place recognition -- while our focus is on traversability analysis to complete the graph.
We do not need to think about loop-closure here because we assume to have access to \ac{GPS} outdoors.

\paragraph{Behavior cloning in robotics}
Behavior cloning is the supervised learning of control policies, where you perform regression on a dataset of state-action pairs.
The data is typically collected by having an expert operating the system.
It has seen a increased interest by the community recently, as a way to tackle manipulation by leveraving large amount of demonstration data, as in RT-2 \cite{rt2}.

Implicit behavior cloning \cite{implicitbc} learns an energy function that has to be minimized post-training to give the action; this allows better modelling of non-smooth policies.

Chang et al. \cite{weidi} explores the use of pre-trained large vision models for behavior cloning and propose a novel keypoint-based behavior cloning approach.
Inspired by this approach, we explore the use a pre-trained large vision model for our behavior cloning controller and our traversability analysis.

\cite{Manderson2020rss} use hindsight relabeling to create a goal-conditioned behavior cloning controller and deploy it on an underwater robot.
We extend this by using a pre-trained vision transformer and long range planning and exploration using topological mapping.

\section{Methods}
Our method consists of multiple interconnected components.
We first present the map representation we chose, a topological map made of a graph of interconnected waypoints in \autoref{sec:topological_map}.
We then present how we decide which waypoints should be connected or not by traversability analysis in \autoref{sec:trav}.
After that, we present our low-level controller for navigating in between waypoints, based on behavior cloning in \autoref{sec:bc}.
We finish by presenting our high-level graph planner and the exploration planner that is built on top of it \autoref{sec:exploration}.

\subsection{Topological mapping and planning}
\label{sec:topological_map}

\begin{figure}[ht]
  \centering
  \includegraphics[width=\columnwidth]{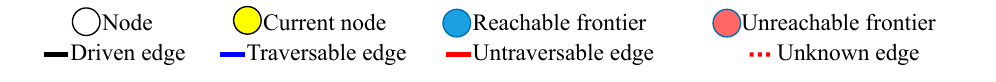} \\
  (a) Initialization of the topological map \\
  \includegraphics[width=\columnwidth]{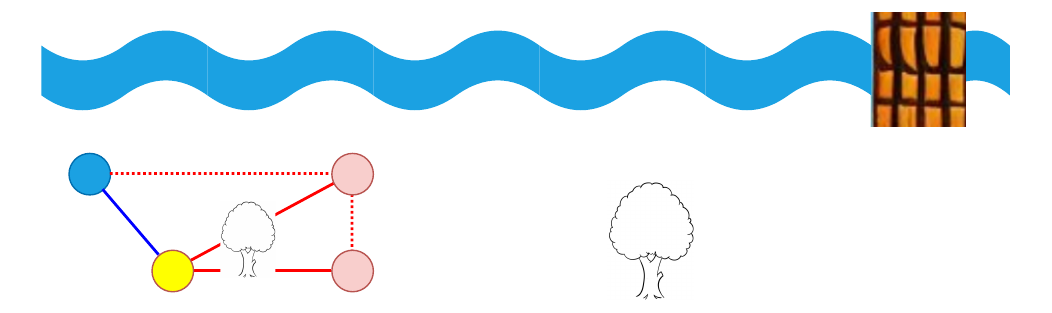} \\
  (b) Start exploration, see river, has to find an alternate path \\
  \includegraphics[width=\columnwidth]{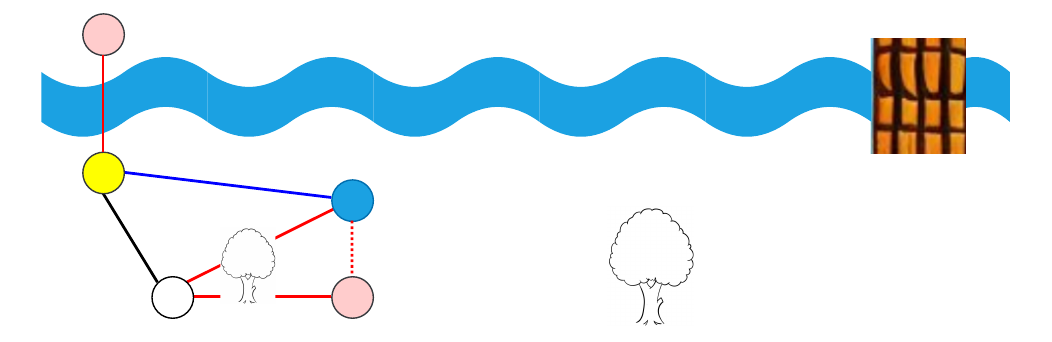} \\
  (c) Exploration complete after find a way to cross the river \\
  \includegraphics[width=\columnwidth]{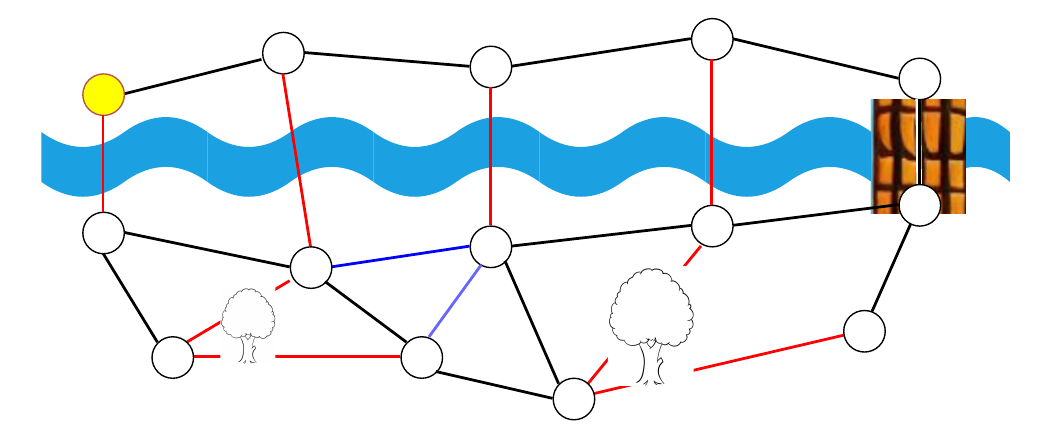}
  \caption{Our topological map allows for a sparse representation of the environments, with nodes being GPS waypoints and edges representing reachability between neighboring nodes, from a traversability point of view.
    In (a), we first initialize the map by computing frontiers and if the path towards them is traversable (\autoref{sec:trav}).
    The closest reachable frontier is selected, and the BC controller goes to its position (\autoref{sec:bc}).
    In (b), after the robot has reached its target node, we recompute the frontiers and select the next one.
    This process is repeated until no reachable frontier nodes are left in the desired zone of exploration, as shown in (c).}
  \label{fig:topological_map}
\end{figure}

We assume a global localization solution with $\approx$ 1 meter of precision is available to use.
For our problem, a topological map is defined as $\mathcal{M} = (\mathcal{N}, \mathcal{E})$ where a node $N \in \mathcal{N}$ contains:
\begin{enumerate}
  \item $N.pose$ the pose in global frame associated with the node
  \item $N.images$: images taken from the robot's point of view and that position, from its left, front and right camera in our case.
\end{enumerate}
Because we do require the global pose of the robot in space, our map could be qualified as a metric/topological hybrid.
The important part is that no terrain modeling is required, and no 3D sensor such as lidar is needed.

We also have unvisited frontier nodes, for the purpose of the exploration planner described in \autoref{sec:exploration}.
These nodes contain no images, only a 3D point coordinate.

An edge $E \in \mathcal{E}$ is either traversable, untraversable or unknown, which we will define in \autoref{sec:trav}.
Every $d$ meter traveled by the robot, a new node is created and added to the map and connected to the previous node with a traversable edge.
Nodes are connected to all of their neighbors within a radius of $r$, larger than $d$ using a $k$-d tree.
The edge type between the neighbors is determined by the method of \autoref{sec:trav}.
$d$ and $r$ are chosen depending on the terrain difficulty and the size of the robot.
The denser the obstacles, the smaller those values need to be.
Automatic and adaptive selection of $d$ and $r$ is left for future work.

\subsection{Traversability analysis}
\label{sec:trav}
In this section, we will define our notion of traversability and outline how we learn to predict the existence of a traversable path between two nodes in our map.

For the purposes of this work, a pixel in an image is considered traversable if the robot can safely be at the visible location imaged in this pixel.
We then extend this definition to define image-level traversability, that is if there exists a traverable path towards a goal that is fully within the image.

We then extend this definition to nodes in our map.

\noindent
\textbf{Definition 1: Image-level traversability}
\begin{align*}
  Trav(I, g) = \exists & \text{ path } \tau(t), t \in [0, T]                                                 \\
                       & : \tau(T) = g \text{ and } P_I\tau(t) \in I \hspace{3pt} \forall t                  \\
                       & \hspace{5pt} \text{ and } P_I\tau(t) \text{ is traversable } \hspace{3pt} \forall t
\end{align*}
Where $I$ is an image from the starting node,
$g$ is the target node's position in $\mathbb{R}^3$ in global frame,
$\tau$ is a trajectory in $\mathbb{R}^3$,
$P_I$ is $I$'s projection matrix, mapping from 3D coordinates to pixel coordinates.

Here we focus on the existence of a visible, safe path to the goal, with the safety determined by a human expert during the learning phase.
The low-level controller is responsible for determining and executing the path towards the goal.

From this image-level traversability, we can define edge-level traversability:

\noindent
\textbf{Definition 2: Edge-level traversability}
$Trav(A, B)$ is true if $\exists I \in A.images : Trav(I, \mathcal{T}_B^A)$ or $\exists I \in B.images : Trav(I, \mathcal{T}_A^B)$

The traversability analysis network $Trav_\theta$ estimates images-level traversability through supervised learning, with edge-level traversability being determined by logic after that.
Given an image and a target location, the output of the network is whether that target is reachable according to Definition 1.
The image encoder has a pre-trained DINOv2 \cite{oquab2024dinov2} backbone for data-efficiency purposes, outputting 384-dimensional features for 256 images patches.
We use three learnable convolutional layers with a filter size of 1, effectively an MLP for each patch, to reduce the dimensionality to 10 for each of the 256 patches.
This is flattened and concatenated to the 3D goal in the local camera frame, which is then fed into a fully connected neural network which outputs two class logits for traversability.
\begin{figure*}
  \centering
  \includegraphics[width=\textwidth]{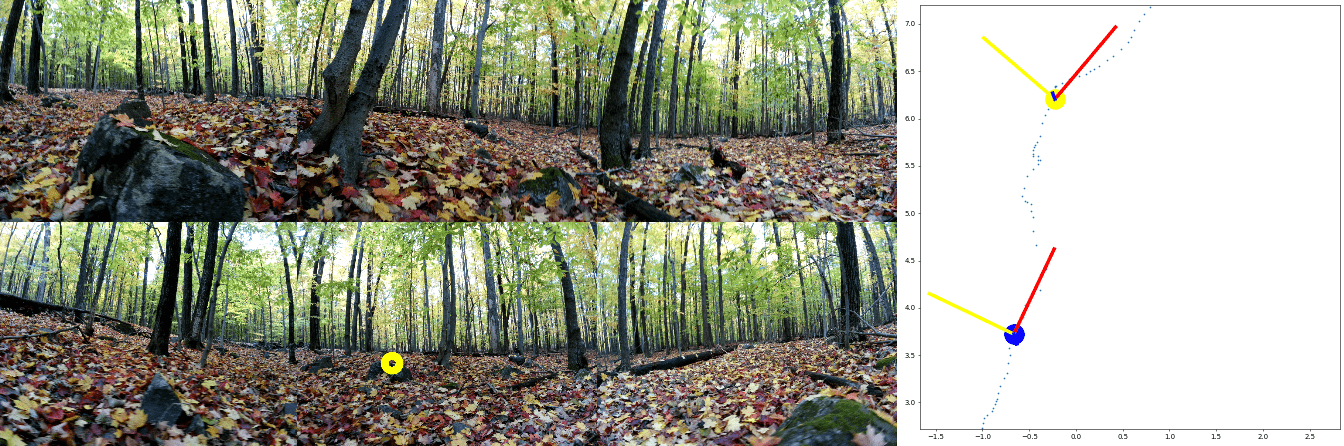}
  \caption{Here the traversability labeling user interface is shown.
    On the left, we have the two panoramas taken from the two different positions.
    When one node is visible from the other, its position is projected in the panorama of the other node.
    On the right, the user has a more global view of the two nodes' position, helping relate the nodes' position to each other.
    In this example, the yellow node is visible from the blue node.
    Here there are no visible obstacles or other visible problems preventing the robot from reaching the yellow node from the blue node.
    This example is thus labeled as traversable by the labeller.}
  \label{fig:labeling}
\end{figure*}

There is also another edge type beyond untraversable and traversable: unknown.
Indeed we cannot determine the traversability $A$ to $B$ if $A$ is not visible from $B$ and vice-versa.
Furthermore, it is impossible to determine the traversability of an edge between two frontier nodes as we have no visual data to do so.

The traversability analysis network is first trained in a supervised way, with labeling done by humans using the interface shown in \autoref{fig:labeling}.
Given the knowledge about the ability of our robot to go in between nodes, we then need a control policy which will navigate on that traversable edge.

\subsection{Goal-conditioned behavior cloning controller}
\label{sec:bc}
To drive the robot from node to node, we need a controller to give linear and angular velocity input to the robot to reach intermediate goals given by the high-level planner.
Smaller obstacles and rough terrain make this task difficult.
Although the higher level traversability model establishes the existence of a feasible path, it does not predict how to get to your goal.

The requirements for the model are as such: it should take as input the current context of the robot (images, velocity, past commands), a visible, nearby goal, and output a linear velocity and turning rate.
First and foremost, collecting demonstration data is simple in our use-case; simply drive the robot around in diverse environments and scenarios.
In cases where this is feasible, behavior cloning has been shown to be a reliable way to learn a control policy in complex scenarios \cite{chi2023diffusion,implicitbc}, although this is the first work to our knowledge to apply this to off-trail driving.

For this problem, we have the action space $\mathcal{A} = [0 \text{ m/s}, 1 \text{ m/s}] \times [0 \text{ rad/s}, 1 \text{ rad/s}]$.
Our policy
\begin{equation*}
  \pi_\theta(o_{t-h+1:t}, a_{t-h+1:t-1}, g) = a_{t:t+\ell}
\end{equation*}
takes as inputs the history of length $h$ for past observations (left, front and right images as well as velocities), past robot inputs and the target goal, in a hindsight relabeling manner \cite{Manderson2020rss}.
It outputs an action \textit{sequence} of length $\ell$, of which only the first action $a_t$ is executed \cite{chi2023diffusion}, similarly to how model-predictive control works.

The objective $\mathcal{L}$ we're minimizing, given the trajectory $\tau(t) \in \mathbb{R}^3$ in global frame, is
\begin{align*}
  \mathcal{L} = \mathbf{E}_{t \in 1:T} \mathbf{E}_{k \in 1:l} || & \pi_\theta(o_{t-h+1:t}, a_{t-h+1:t-1}, \mathcal{T}_{R_t}^G \tau(t \hspace{-2pt}+ \hspace{-2pt}k)) - \\
                                                                 & \pi_{e}(o_{t-h+1:t}, a_{t-h+1:t-1}, \mathcal{T}_{R_t}^G \tau(t \hspace{-2pt}+\hspace{-2pt} k))||^2
\end{align*}
Where $k$ samples different goals at different horizon in the future, with the maximum sampling time for the goal being $l$.
We transform the sampled goal in the robot's local frame at time $t$, using the robot pose $R_t$.

The neural network architecture for $\pi_\theta$ consists of a shared encoder for all images from the left, front and right cameras.
The image encoder has the same DINOv2 and dimension reduction as $Trav_\theta$ in \autoref{sec:trav}, but is trained separately.
This is followed by a fully-connected multilayer perceptron to fuse the output of the encoder for all images, the velocities (delta poses represented by 3x4 matrices), and the goal in local frame.

Now that we have a method for navigating in between nodes in the graph were the path is visible, we need a higher lever planner to allow longer-horizon reasoning.

\subsection{Exploration planning}
\label{sec:exploration}
\begin{figure}[ht]
  \centering
  \includegraphics[width=\columnwidth]{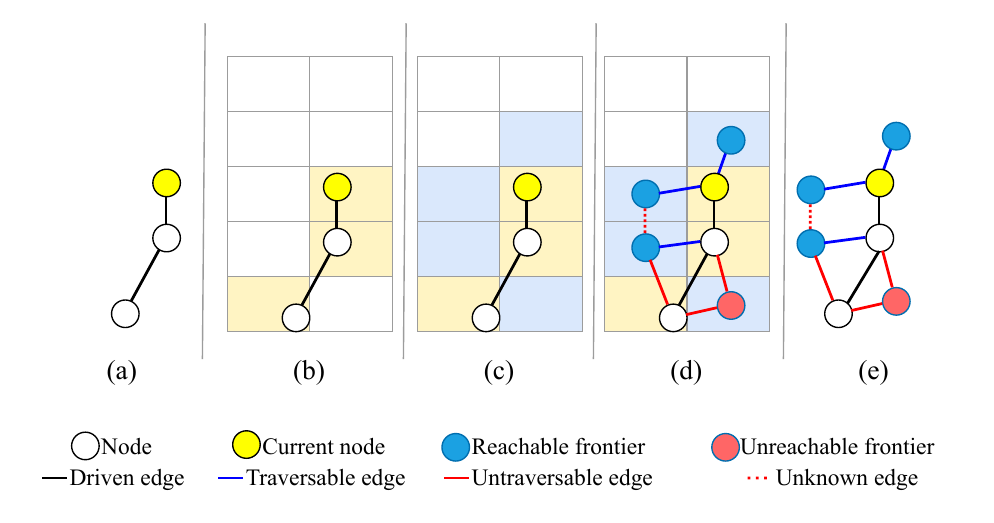}
  \caption{An illustration of the frontier creation algorithm.
    (a) starting map
    (b) creation of the occupancy grid
    (c) computing the frontier cells in the grid
    (d) creating the frontier nodes from the cells and adding them to the map
    (e) final result.
    The graph, containing traversability information, can then be used to plan further exploration.}
  \label{fig:exploration_alg}
\end{figure}

To explore the environment and build a traversability-aware topological map, we use a frontier-based exploration algorithm.
We define, for this work, a frontier as being an unexplored node $N$ without $N.images$ that is close to a visited node in our map.
More specifically, we create an occupancy grid of $R \times R$ cells of size $\lambda$ around the starting point.
A cell is occupied if a visited node falls within it, and a frontier if a neighbour cell is occupied.
We can then create a frontier node at the center of the frontier cell and add it to our graph map.
This process is illustrated in \autoref{fig:exploration_alg}.

Now that we can create frontier nodes as exploration candidates, we need to plan to explore them judisciously.
We use here a heuristic method to estimate which frontier is the closest time-wise to the current robot position.
Rather than simply using the euclidean distance, we use the time it would take to turn the robot towards the frontier and then drive straight to it using a basic kinematics model of our skid-steered robot.
We found this kinematics-based heuristic to be important, as using the euclidian distance led to a lot of turning around which wasted time and energy.

After selecting the frontier, we plan to it using A\textsuperscript{*} along traversable edges, which gives us a sequence of waypoints in our graph.
We take only the first waypoint of this plan, and give it as a goal to our behavior cloning controller.
When that first waypoint is reached, we compute the frontiers using the method of \autoref{fig:exploration_alg} again, select a frontier again and re-plan using A\textsuperscript{*}.
This re-planning approach allows us to:
\begin{enumerate}
  \item change the target frontier if a better candidate arises,
  \item use a better path towards the target frontier if one is uncovered during navigation
  \item replan in the case of traversability estimation error detected from BC goal timeout or human intervention
\end{enumerate}
This process is repeated until no reachable frontier nodes are left in the map.

\section{Experimental setup}
\subsection{Robot platform}
The robot is a skid-steered Clearpath Husky, which has the required durability for navigating off-road terrain, sustaining collisions and various shocks and scratches from branches.
However, it is quite limited in mobility for our application: it cannot climb over large fallen trees, cannot go in water like small streams or mud.
Our method thus adapts to this limited mobility and plans paths that are aware of the capability of our platform.
The robot is equipped with three OAK-D wide-angle lens cameras, one on its left, front and right.
It has Microstrain 3DM-GX5-25 \ac{IMU} and a \acf{RTK} \ac{GPS} system.
For compute, the robot is equipped with an Intel NUC and a Nvidia Jetson Orin for neural-network inference.
It is important to note that our method is generic and can be used for other robot platforms, as long as it can be piloted by a human to give demonstrations.

\subsection{Experimental zone}
The real-world experiments were conducted at Maisonneuve Park, a municipal park in Montreal, and Gault Nature Reserve, a mountain and nature reserve that is owned by McGill University.
The forest composition and terrain type is varied, with mature maples, beech and oaks dominating.
Large fallen trees are present everywhere and the robot must plan around them.
Other obstacles present include large rocks, cliffs, hills that are too steep, ponds or streams, as seen in \autoref{fig:trav_map}.

\subsection{Training and deployment methodology}
Here we describe the training methodology which makes all these previous element come together, from manual data collection and labeling, to semi-autonomous deployments and finally to test-time deployments in unseen terrain.
All data was collected at 5 Hz, which was the rate at which we executed the behavior cloning policy.
Our different parameters during training were as such: $r$ was 3 m, $d$ was 1 m, $h$ was 5, $l$ was 25, $\ell$ was 5 and $\lambda$ was 2 m.

\textbf{Step 1: Manual data collection}
Before labeling any traversability data and learning any behavior cloning policy, a human needs to drive the robot manually.
This is done in different areas, trying to maximize terrain diversity and driving safely in different scenarios.
In this phase, 8.5 kilometers were driven manually in 24 hours of total driving time.
The process spanned late Summer 2023, Fall 2023 (including some snowy days) and Spring 2024.

\textbf{Step 2: Manual traversability annotation}
We then take the topological maps created during step 1 and manually annotate the edges using the interface shown in \autoref{fig:labeling}.
We compute the frontiers of the maps of step 1, and annotate the edges leading to those frontiers.
This leads to more untraversable examples being labeled.
In this phase,  3855 edges were manually labeled for traversability.

\textbf{Step 3: Neural network training}
We then train both the behavior cloning policy $\pi_\theta$ and the traversability network $Trav_\theta$.
This is done offline on a compute cluster, not directly on the robot.
While online learning of traversability has been done \cite{frey2023fast}, we favor offline learning to leave room to execute both the behavior cloning and traversability network on the robot.
Learning offline also allows for the use of bigger models and learning with larger datasets.

\textbf{Step 4: Semi-autonomous deployment and data collection}
Now that we have both models trained, we can deploy them on the robot and use them to collect more data.
We semi-autonomously explore multiple areas, using the algorithm of \autoref{sec:exploration}, with human intervention when necessary.
The data is added to the behavior cloning data through hindsight relabeling, leading to better performance.
We also do the manual traversability annotation with the data collected in this phase.
\textit{We can return to step 3 and retrain as many time as needed.}
In this phase, 2 kilometers were additionally collected for behavior cloning, in 10 hours during Summer 2024.
and 1679 edges were added to the dataset.

\textbf{Step 5: Test deployment}
We can finally go to an unseen forest site and deploy the exploration algorithm to map new environments.
In practice, this is still done with a human supervisor for robot safety reasons.
Interventions by the supervisor are reported on, and the data is not reused for training.
Here, as seen in \autoref{fig:trav_map}, two zones at the Gault Nature Reserve were autonomously explored and mapped.
These zones were geographically distinct from the training data.
Site A was next to the training zone, but did not overlap, while site B was a few kilometers away next to a lake and had a different composition than seen during training.
Indeed, being at the foot of a mountain, there was a steep incline, and the forest was more mature with very large fallen trees littering the ground, making navigation challenging.

\section{Results}
\begin{figure*}[ht]
  \centering
  {\small \textbf{Test A}} \\
  \vspace{3pt}
  \includegraphics[width=0.48\textwidth]{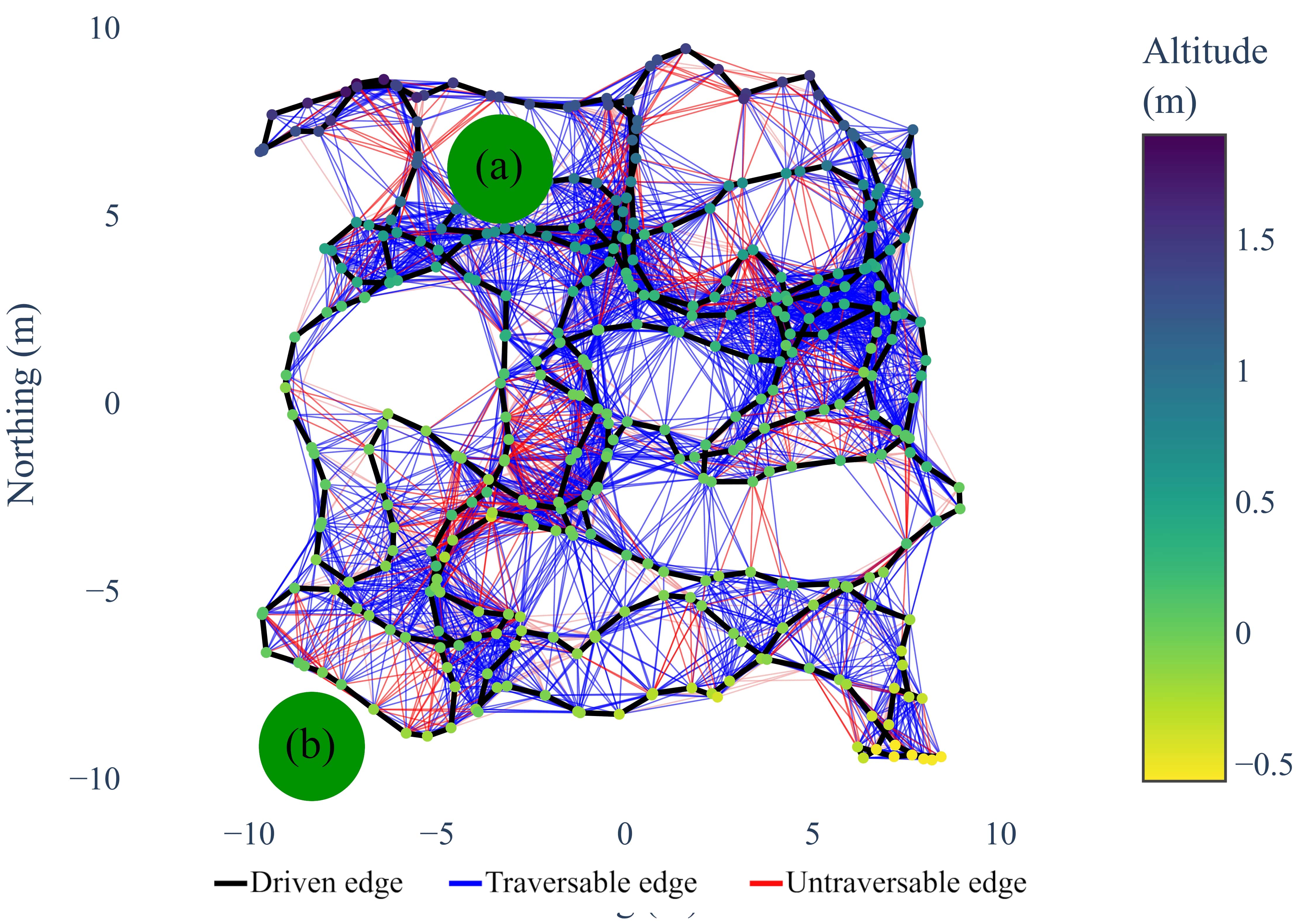}
  \includegraphics[width=0.24\textwidth]{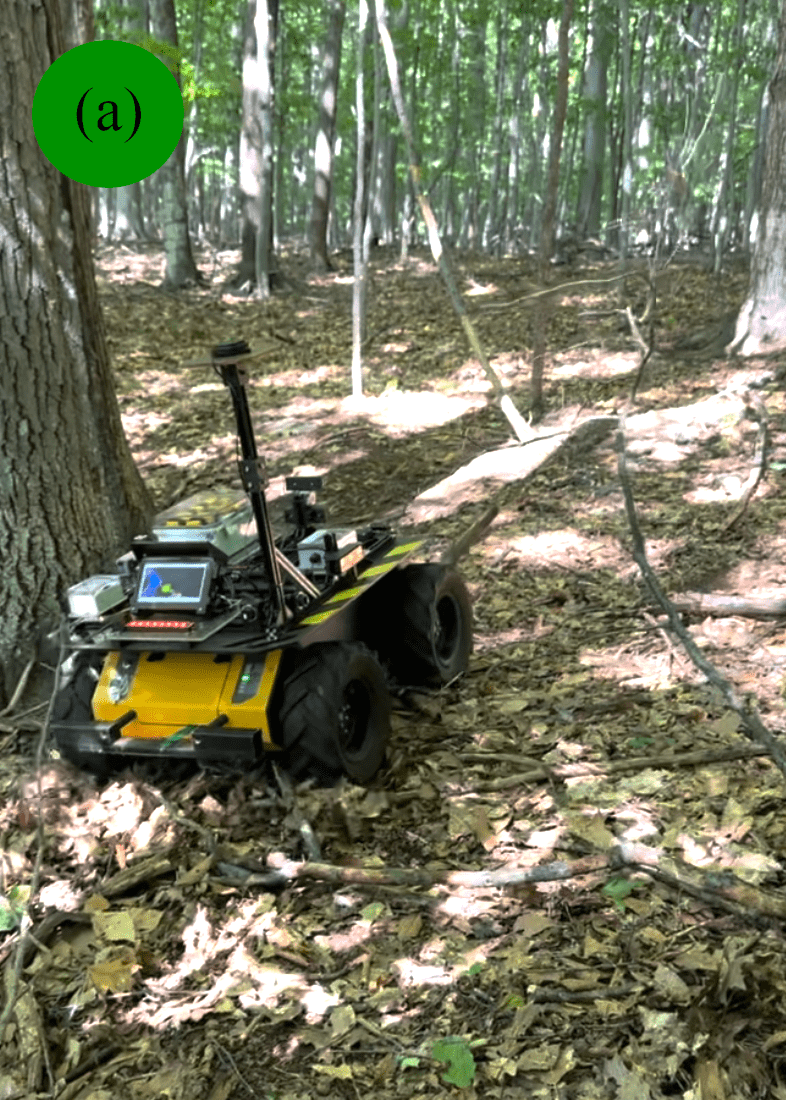}
  \includegraphics[width=0.24\textwidth]{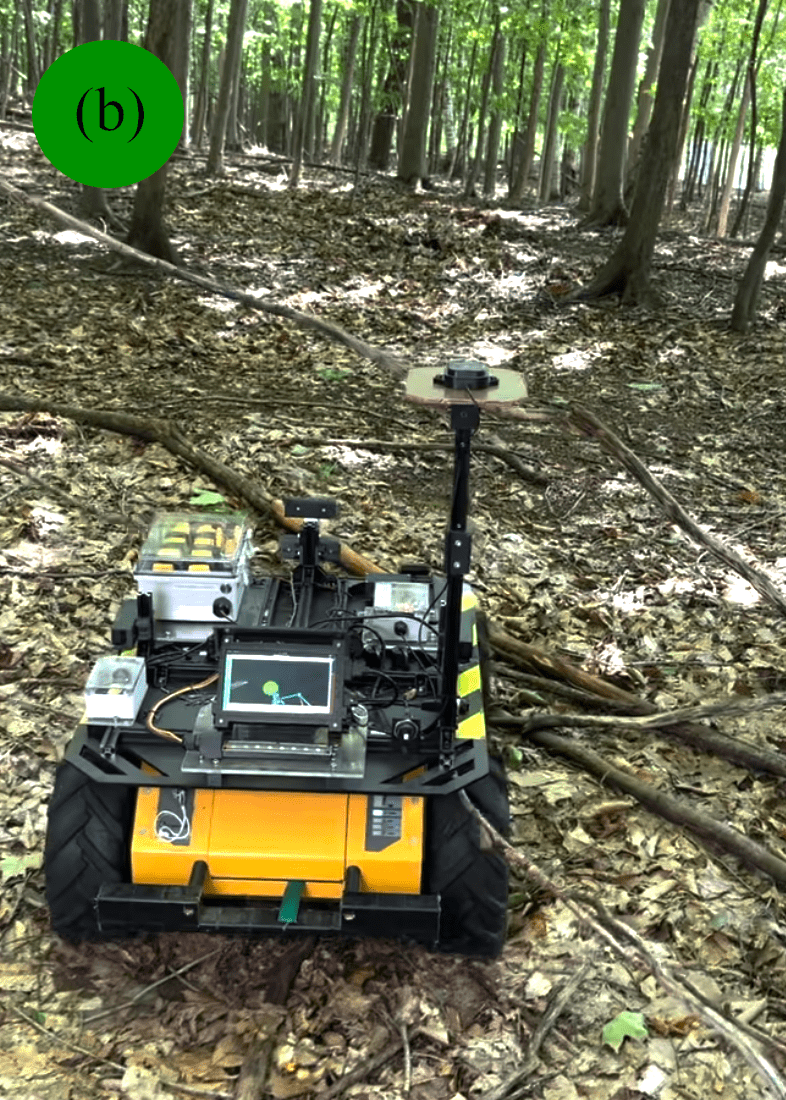}
  \\
  {\small \textbf{Test B}} \\
  \vspace{3pt}
  \includegraphics[width=0.48\textwidth]{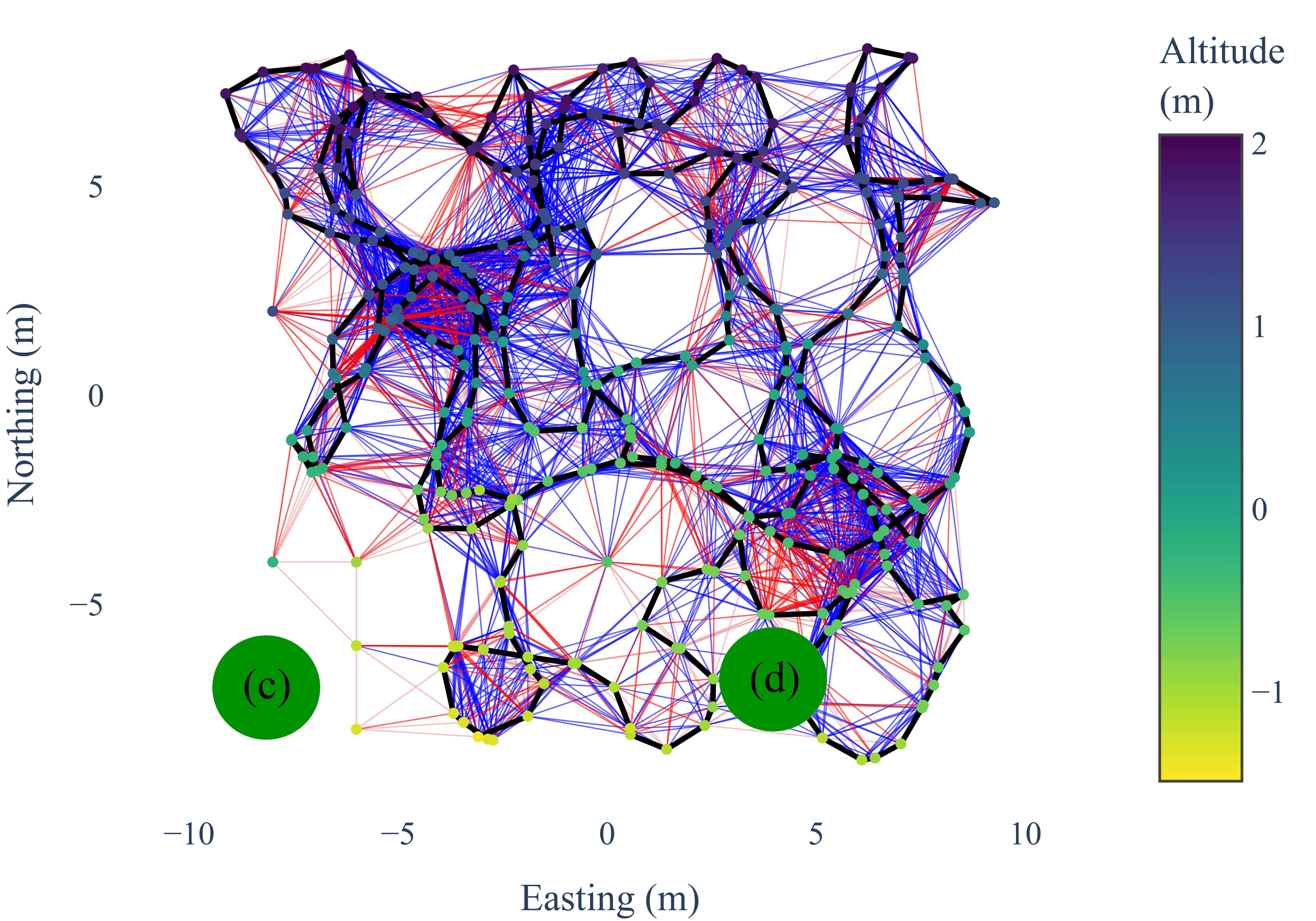}
  \includegraphics[width=0.24\textwidth]{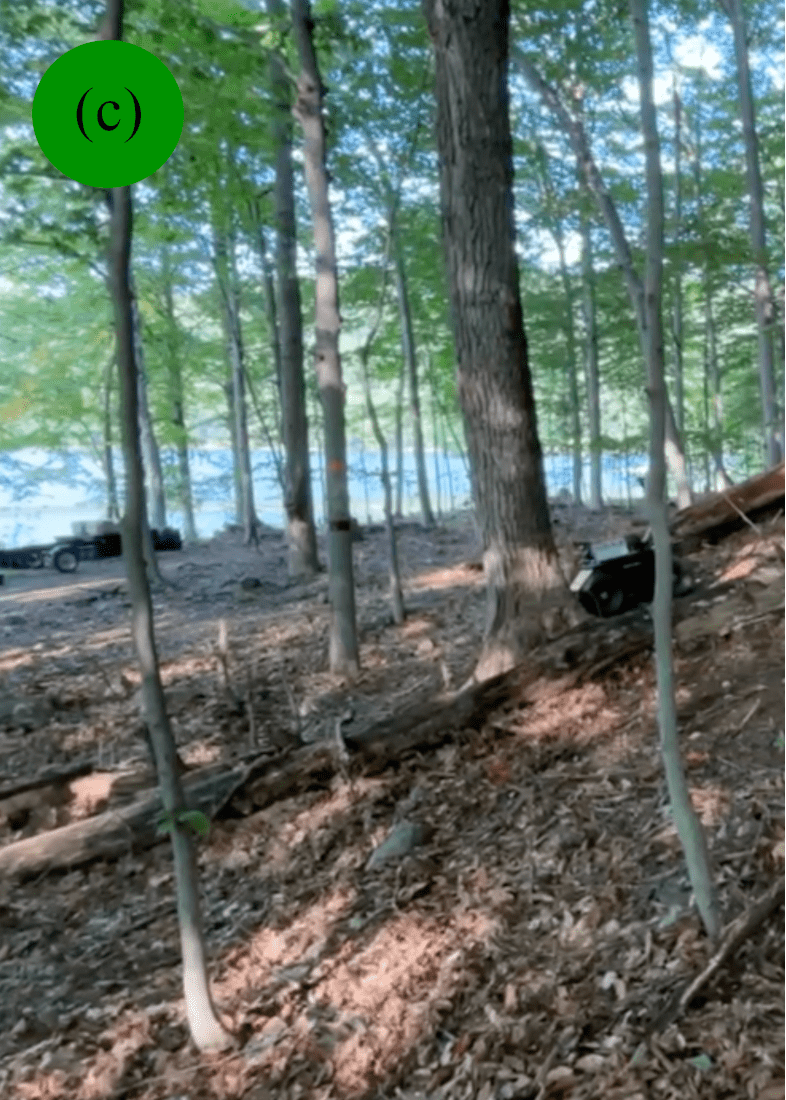}
  \includegraphics[width=0.24\textwidth]{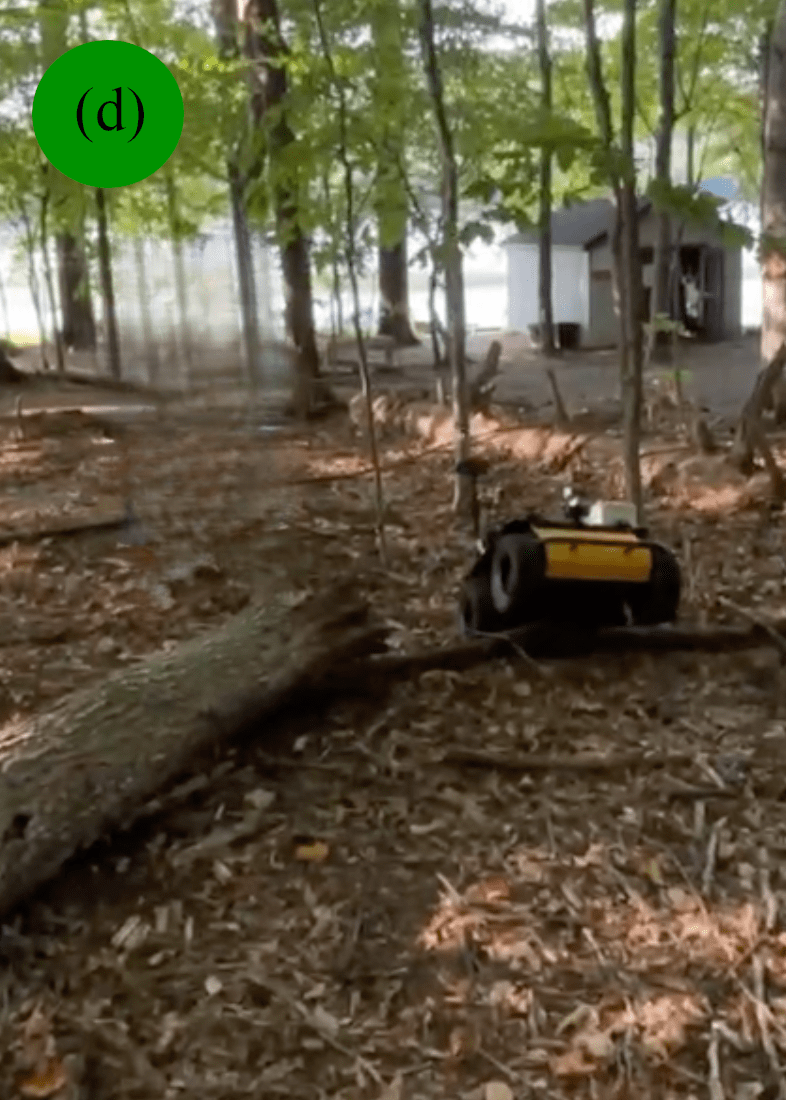}
  \caption{
    Topological maps generated using our autonomous exploration process on two unseen environment.
    Nodes in the graph are color-coded as a function of the terrain height, highlighting the rough and hilly nature of the terrain.
    In testing zone A, (a) demonstrate how the robot found a path in between a large tree and a pile of branches that would have been difficult to traverse.
    In (b) another pile of branches is perceived as both untraversable and traversable, highlighting the viewpoint sensitivity of our approach.
    Indeed, there is a large branch high off the ground which could block the robot depending how the robot is approaching the pile.
    Now in testing zone B, (c) shows multiple fallen tree completely block a zone from being explored.
    However, in (d), we see a large fallen tree correctly detected as untraversable, which the robot is able to find a path around and keep exploring.
  }
  \label{fig:trav_map}
\end{figure*}

We now present the results of the autonomous exploration of both test site A and B.
Both topological maps built using our approach are shown in \autoref{fig:trav_map}.
An analysis of the traversability network results is presented in \autoref{sec:trav_results}, followed by a few comments on the performance of the behavior cloning controller in \autoref{sec:bc_results}.
We finally comment on the exploration and overall system performance in \autoref{sec:expl_results}.

\begin{figure}[ht]
  \centering
  \includegraphics[width=1.0\columnwidth]{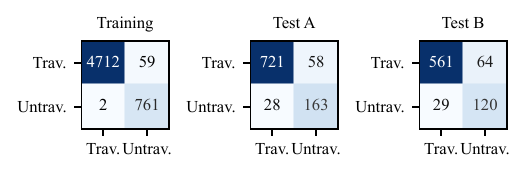}
  \caption{
    Confusion matrices for the training set, as well as manually labeled data from the two test runs.
    The rows represent the ground truth, while the columns represent the model's prediction.
  }
  \label{fig:trav_roc}
\end{figure}

\subsection{Traversability}
\label{sec:trav_results}
After the test deployment, for evaluation purposes, we manually labeled the traversability dataset of test site A and B as we had done with the training data.
The comparison of predicted traversability and human-labeled traversability is shown in \autoref{fig:trav_roc};

However, classification error does not tell the whole story as not all mistakes are equal.
False negatives (the model outputting untraversable when it is traversable) were of little consequence in our two test explorations, as in practice plenty of neighboring edges were correctly classified as traversable.
The planner could then use these edges to keep exploring.
Even false positive, which could be more problematic, were often not cause for intervention because the robot would find a path to the goal, even though it was not visible from the node of departure.
Ultimately, 9 interventions were caused by going through untraversable edge incorrectly classified, leading the robot to dangerous terrain.


\subsection{BC controller}
\label{sec:bc_results}
In total, 11 interventions were caused by the BC controller, mostly because the robot hit small trees rather than going around them.
In step 4 of our deployment, we noticed a steady improvement in performance as more data was collected.
We thus hypothesise that the controller will keep improving in the future with more deployments.

\subsection{Planning and full system}
\label{sec:expl_results}
The planner led to a successful exploration, visiting every reachable cell of 2 m $\times$ 2 m of the two test zones, with 700 m of travel.
In total, there were 25 interventions.
The GPS antenna was above one meter above the ground, and not visible from the robot's camera.
While traversability decision were made with this in mind, the behavior cloning controller, with a memory of 5 frames or 1 s, could "forget" that there was something that could block the antenna and needed to be carefully avoided.
This caused 5 interventions, on top of the ones caused by the controller or traversability analysis.
This could be fixed by adding another camera monitoring the clearance of the antenna.

\section{Conclusion}
\label{sec:conclusion}
This paper presents an algorithmic framework enabling navigation in off-trail environments, pushing the boundaries of what kind of terrain a robot can explore autonomously.
This is achieved without the use of dense map building and fined-grained path-planning, but rather a sparse, traversability-aware topological map and a simple behavior cloning controller.
Combined with our frontier-based exploration approach that is continously replanning in the face of uncertain terrain, this enabled the autonomous exploration of two unseen 400 m\textsuperscript{2} forest zones.
Future work will focus on vision-based frontier selection, and actively collecting traversability data to reduce labeling needs.

\bibliographystyle{IEEEtran}
\bibliography{IEEEabrv, references.bib}

\end{document}